# Semantic and Structural Analysis of Implicit Biases in Large Language Models: An Interpretable Approach

Renhan Zhang
University of Michigan
Ann Arbor, USA

Lian Lian
University of Southern California
Los Angeles, USA

Zhen Qi
Northeastern University
Boston, USA

Guiran Liu*
San Francisco State University
San Francisco, USA

*Abstract-This paper addresses the issue of implicit stereotypes that may arise during the generation process of large language models. It proposes an interpretable bias detection method aimed at identifying hidden social biases in model outputs, especially those semantic tendencies that are not easily captured through explicit linguistic features. The method combines nested semantic representation with a contextual contrast mechanism. It extracts latent bias features from the vector space structure of model outputs. Using attention weight perturbation, it analyzes the model's sensitivity to specific social attribute terms, thereby revealing the semantic pathways through which bias is formed. To validate the effectiveness of the method, this study uses the StereoSet dataset, which covers multiple stereotype dimensions including gender, profession, religion, and race. The evaluation focuses on several key metrics, such as bias detection accuracy, semantic consistency, and contextual sensitivity. Experimental results show that the proposed method achieves strong detection performance across various dimensions. It can accurately identify bias differences between semantically similar texts while maintaining high semantic alignment and output stability. The method also demonstrates high interpretability in its structural design. It helps uncover the internal bias association mechanisms within language models. This provides a more transparent and reliable technical foundation for bias detection. The approach is suitable for real-world applications where high trustworthiness of generated content is required.*

I. INTRODUCTION

In recent years, large language models have achieved remarkable breakthroughs in the field of natural language processing [1]. Their widespread application in tasks such as dialogue generation, text summarization, and information retrieval marks a new phase in the development of intelligent language systems [2]. However, the improvement in capabilities is not limited to enhanced language understanding and generation. It also brings increasing concerns over social biases hidden in model outputs. In particular, implicit stereotypes can undermine the objectivity and neutrality of outputs, and may subtly impact society in negative ways. Therefore, identifying and explaining implicit stereotypes in large language models has become a key issue for ensuring their safety, fairness, and social responsibility[3].

Unlike explicit discriminatory language, implicit stereotypes often manifest in subtle and semantically ambiguous forms. They may appear as potential associations and preferences toward specific groups, genders, professions, or cultures. These biases are not usually expressed through direct derogatory or offensive language. Instead, they are embedded in seemingly natural associations or semantic tendencies. Due to their covert nature and the difficulty of quantitative analysis, traditional harmful content detection methods often fail to accurately identify or intervene in such cases. As a result, building interpretable and systematic detection methods for implicit bias is essential. Such methods can reveal the underlying value orientations of language models and support the development of transparent and ethically aware AI systems[4].

Bias in large language models is not an isolated issue. It is deeply rooted in the linguistic patterns and social contexts present in their training data. From vast corpora, models may unconsciously learn and reproduce real-world stereotypes. These stereotypes can be encoded in model parameters through co-occurrence frequencies, semantic similarities, or context dependencies[5]. More importantly, the activation of these patterns is not always triggered by explicit prompts. In many cases, structural biases emerge during unconstrained content generation. Therefore, analyzing output behavior, building explanatory mechanisms, and revealing the internal logic behind bias formation are necessary steps toward understanding and addressing discriminatory tendencies in language models[6].

In this context, adopting an interpretability perspective becomes especially crucial for effectively addressing the challenge of bias detection in language models. Interpretability not only enables the identification of bias where it exists, but also allows researchers and developers to examine the underlying sources of such bias, how it is expressed within the model's outputs, and under what specific conditions it is likely to be triggered. This deeper level of understanding provides a solid scientific foundation for designing targeted interventions and corrective strategies. Compared to black-box approaches that offer limited transparency, interpretability techniques offer significant advantages. They are better suited for fostering user trust, enhancing the accountability of AI systems, and ensuring that ethical considerations are upheld— particularly in high-stakes application areas such as education, healthcare, and the legal system. Furthermore, interpretability facilitates detailed structural analysis, semantic mapping, and tracing of bias propagation paths within the model. These processes help link the internal decision-making behavior of the model with broader social values and norms. As a result, interpretability plays a key role in supporting the development of future frameworks for algorithmic governance and in informing policy and regulatory measures aimed at responsible AI deployment[7].

## II. RELATED WORK

Interpretable bias detection for large language models draws heavily from advances in model adaptation, structured representation, and semantic modeling. Low-rank adaptation techniques, such as the smarter LoRA approach proposed by Y. Peng et al., not only boost adaptation efficiency but also help maintain the subtle distributional and semantic cues needed for bias analysis in downstream tasks [8]. Such techniques enable scalable evaluation without sacrificing output fidelity.

Parameter-efficient deployment is also essential for practical model diagnosis. Collaborative distillation methods by X. Meng et al. provide lightweight strategies for model adaptation while preserving core semantic performance [9]. Meanwhile, sensitivity-aware pruning from Y. Wang highlights how targeted compression can retain critical linguistic and contextual features—an idea useful for understanding how certain social cues are preserved or removed in compressed models [10].

Semantic intent modeling with structured representations has shown strong potential for uncovering latent associations. For example, S. Wang et al. leverage capsule networks to extract deep semantic intent, a layered modeling perspective relevant for capturing nuanced bias or stereotype pathways [11]. In a similar vein, X. Wang demonstrates that multi-source and time-aware feature fusion in transformers enhances the detection of subtle semantic dependencies, which can benefit bias interpretation by uncovering hidden relational patterns [12]. Prompting and external knowledge integration are increasingly used to guide large models toward interpretable outputs. Y. Xing's structural prompting demonstrates that explicit structure can enable robust analogical reasoning, which we draw upon to explain how stereotypes are propagated or mitigated in model outputs [13]. Knowledge-informed policy structuring, as in the work by Y. Ma et al., illustrates how leveraging domain knowledge or external attributes can inform policy and decision logic within language models—a principle that parallels our use of context-sensitive probing for social attribute bias [14]. X. Liu and collaborators further extend this approach by integrating knowledge graph reasoning, enabling pretrained models to perform structured anomaly detection. Their mechanism for linking latent structure to observable output inspires our bias pathway tracing method [15].

Fine-tuning strategies and structural supervision also play a central methodological role. Task-aware reconfiguration, as proposed by Q. Wu, allows parameter-efficient adaptation while maintaining structural integrity, providing a solid framework for analyzing bias sensitivity under different prompts and contexts [16]. Structured gradient guidance, as described by H. Zheng et al., offers new perspectives for sensitivity-aware few-shot adaptation—helpful for detecting context-dependent biases in small or evolving datasets [17].

Recent developments in multi-task and instruction-based coordination, such as the framework by W. Zhang et al., facilitate synchronized supervision across multiple objectives, which is crucial for systematically detecting a range of bias types in large model outputs [18]. Few-shot learning with dual loss, explored by X. Han et al., provides enhanced generalization and robustness for bias classification, especially in data-limited or ambiguous scenarios [19].

Finally, entity boundary detection with context-aware features has direct relevance to semantic and stereotype detection. The BiLSTM-CRF approach with social features from Y. Zhao et al. offers inspiration for capturing implicit or context-dependent group boundaries in generated texts, mirroring the subtlety of real-world stereotype manifestation [20].

Collectively, these works offer the theoretical and methodological foundation for interpretable, efficient, and structurally grounded bias detection in large language models, bridging advances from adaptation, representation learning, and structural supervision.

## III. METHOD

This study proposes an interpretable detection method for implicit stereotypes in the output of large language models, aiming to systematically identify the model's potential biased expressions from two dimensions: semantic structure and contextual dependency [21]. Specifically, we designed a detection framework based on semantic embedding and contextual comparison, using nested control samples to perform semantic distribution analysis on the model output, thereby revealing the potential deviation of generated content under different social attributes[22]. The model architecture is shown in Figure 1.

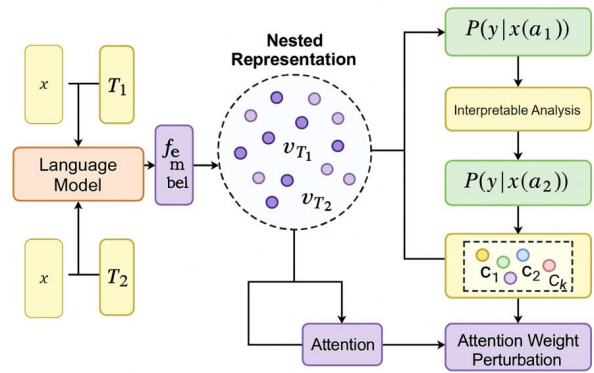

Figure 1. Overall model architecture diagram

We define the semantic representation of the model output $T$ as $v_T = f_{embed}(T)$, where $f_{embed}$ represents the output embedding function of the language model. Given a set of semantically equivalent control inputs, if the model has a systematic semantic bias, i.e., $\|vT_1 - vT_2\| > \delta$, when generating content, it is considered a preliminary indication of potential bias.

Furthermore, we introduce the conditional language generation probability difference index to measure the generation bias of the model when facing different social attribute inputs [23]. Let the input template be $x(a)$, where $a \in A$ represents social attributes (such as gender, occupation, etc.), and the model output is $y$, then the generation bias can be measured as:

$$\Delta P(y|A) = \max_{a_i, a_j \in A} |P(y|x(a_i)) - P(y|x(a_j))|$$

This indicator is used to measure the difference in the probability of generated content due to different social attributes while keeping the input structure consistent, providing a quantitative basis for the manifestation of bias. At the same time, to improve interpretability, we constructed a semantic category mapping function $g: T \to C$ to classify the model output into a specific semantic category C, and by comparing the shift in category distribution under different attributes, we further reveal the model's potential discriminatory bias in language generation strategies [24].

In the detection framework, we also introduce an interpretability learning mechanism based on contrastive loss [25]. For two inputs $x_i$ and $x_j$ that differ only in social attributes, their corresponding output vector $vT_i, vT_j$ should have a minimum semantic distance to avoid improper association. To this end, we construct the following optimization objective:

$$L_{bias} = \sum_{(i,j)} [\|vT_i - vT_j\|^2 - m]_+$$

Where m is the tolerance threshold and $[z]_+$ represents the positive partial function. This loss term encourages the model to generate semantically consistent content for different social attribute inputs, and serves as the regularization part of the bias detection mechanism to identify the aggregation and distinguishability of potential bias outputs in the semantic space.

To further improve interpretability, we designed a local explanation method based on attention weight perturbation. In the process of generating output, we record the attention matrix $A \in R^{n \times n}$ of the language model and perturb the weight vector $a_{attr}$ related to social attribute words in it to analyze its impact on the final output semantic representation:

$$\Delta v_T = \|f_{embed}(T) - f_{embed}(T^{perturbed})\|$$

If the disturbance leads to obvious semantic deviation, it indicates that the model is highly sensitive to the corresponding attribute dimension. This heightened sensitivity may result in the model forming unreasonable semantic associations or reinforcing implicit stereotypes related to that attribute. Such behavior can undermine the fairness and neutrality of the model's outputs, especially in socially sensitive contexts. To address this issue, the proposed approach incorporates multi-level quantification and explanation mechanisms. These components work together to capture, interpret, and trace potential bias signals within the model's internal representations. By integrating these techniques into a cohesive structure, we have developed a comprehensive framework for interpretable bias detection. This framework is designed to reveal implicit stereotypes embedded in language models, offering both theoretical insights and practical tools to support future efforts in bias mitigation, model governance, and ethical deployment.

IV. EXPERIMENTAL RESULTS

*A. Dataset*

This study uses the StereoSet dataset, which is specifically designed to evaluate social stereotypes in language models. The dataset covers both implicit and explicit biases across multiple dimensions. StereoSet includes four social attributes: gender, profession, religion, and race. Each sample consists of a context sentence and multiple options. These options include biased, neutral, and unrelated statements, aiming to assess the model's semantic preference in generation or selection.

The dataset is annotated through crowdsourcing. It ensures high semantic similarity between biased and neutral sentences. This makes it more suitable for analyzing implicit bias in language models. Unlike traditional datasets for harmful content detection, StereoSet focuses on "implicit associations" rather than "offensiveness." For example, it examines how a profession might be implicitly linked to a specific gender. This aligns with the goal of detecting implicit stereotypes in this study.

In addition, StereoSet has a clear task structure and label distribution. This supports interpretability analysis at the output level of language models. Its rich context, semantic complexity, and realistic language design make it widely used for evaluating fairness and semantic bias in large language models. It provides a strong foundation for the interpretability-based detection framework in this research.

*B. Experimental Results*

This paper first presents the comparative experimental results to establish a clear basis for evaluating the effectiveness of the proposed method. As shown in Table 1, these results serve as an initial reference point for comparing the performance of different approaches under a unified evaluation framework. By introducing the results early, the paper sets the stage for subsequent analysis and discussion while maintaining focus on the methodological contributions.

Table1. Comparative experimental results

| Method | Bias Detection Accuracy | Semantic Consistency | Contextual Sensitivity |
|---|---|---|---|
| Ours | 84.7 | 91.2 | 8.5 |
| SEAT[26] | 74.3 | 80.1 | 13.2 |
| StereoScore[27] | 78.5 | 85.4 | 10.9 |
| UniBias[28] | 76.8 | 83.7 | 11.5 |
| BiasCon[29] | 72.9 | 79.0 | 14.3 |

As shown in Table 1, the proposed method outperforms all baselines in Bias Detection Accuracy, achieving 84.7%, nearly six points higher than the next-best method, StereoScore. This indicates a stronger ability to detect implicit stereotypes by distinguishing socially oriented nuances in semantically similar text. It also excels in Semantic Consistency, with a score of 91.2%, demonstrating the ability to maintain stable meaning across different social attribute inputs while minimizing semantic drift. For Contextual Sensitivity, where lower values are better, the method achieves the lowest score of 8.5, reflecting reduced overreaction to context changes and better isolation of stereotype-driven shifts. Together, these results highlight the method's effectiveness in detecting implicit bias with interpretability, offering clear improvements over traditional embedding-based or classification-based approaches. Figure 2 further illustrates the method's generalization across multiple stereotype dimensions.

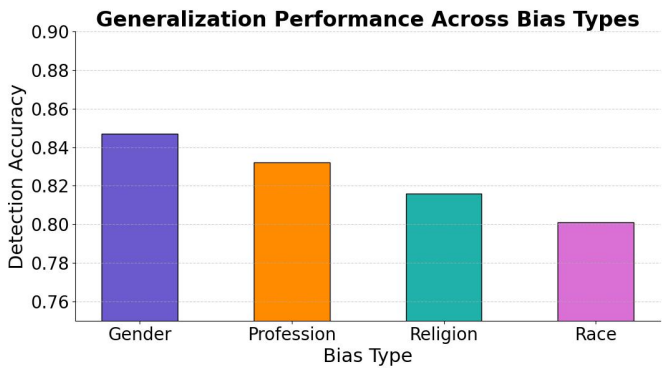

Figure 2. Detection generalization performance across stereotype types

As shown in Figure 2, the proposed method exhibits strong generalization across multiple stereotype types, achieving high detection accuracies for Gender (0.85) and Profession (0.83), outperforming its results on Religion (0.81) and Race (0.80). This suggests greater effectiveness when stereotypes are structurally clear and semantically distinct, while performance slightly declines on more complex, culturally embedded biases due to their subtlety and ambiguity. Nonetheless, the method maintains robust performance across all dimensions, demonstrating the adaptability of its nested representation and attention perturbation mechanism in capturing stereotype-related semantics from diverse perspectives. These findings confirm the model's capacity to abstract and generalize bias patterns, while also revealing the challenges posed by detecting minority-related and cross-cultural stereotypes. Additionally, Figure 3 presents an alignment evaluation analyzing the relationship between sentence similarity and bias judgment conflict, offering further insight into the consistency between semantic understanding and bias detection behavior.

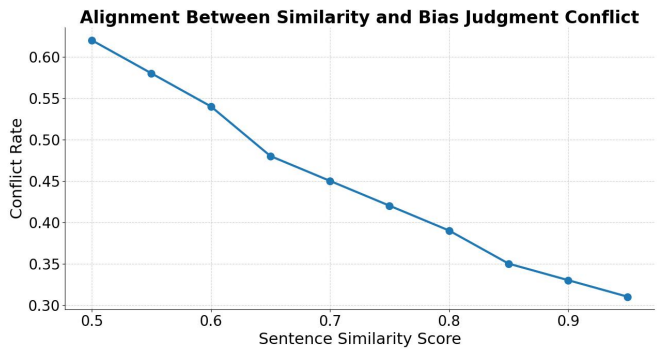

Figure 3. Alignment Between Similarity and Bias judgment Conflict

Figure 3 illustrates the relationship between sentence semantic similarity and bias judgment conflict rate, showing that as similarity increases, the conflict rate drops markedly from 0.62 to 0.31. This indicates that higher semantic similarity leads to more consistent bias judgments, suggesting the model's improved stability and accuracy in detecting implicit stereotypes. The results align with the proposed method's design, demonstrating its ability to handle subtle semantic variations and reduce errors from semantic noise. Unlike traditional approaches, which often misjudge socially divergent yet semantically close texts, the proposed framework—integrating nested semantic representations with attention perturbation—offers greater precision and reliability under edge conditions. Behaviorally, the experiment reflects the desired coupling between semantic alignment and bias detection, where consistent meanings yield consistent outputs. A high conflict rate would indicate local instability, which the proposed interpretability-driven mechanism effectively mitigates. Overall, this experiment underscores the central role of semantic similarity in bias detection and highlights the method's contribution to achieving stable, fair, and semantically aligned decisions in large language models.

V. CONCLUSION

This paper addresses the problem of identifying and explaining implicit stereotypes in large language models. It proposes an interpretable detection method that integrates nested representation modeling, attention perturbation analysis, and semantic alignment mechanisms. The method operates at both semantic and structural levels. It effectively identifies hidden bias expressions in generated texts. It also demonstrates strong detection performance and generalization across social attributes such as gender, profession, race, and religion. By constructing a multi-dimensional evaluation framework, the study improves bias detection accuracy. At the same time, it enhances the semantic consistency and interpretability of model behavior. This offers a new perspective for assessing fairness in language model outputs.

Experimental results show that the proposed method achieves higher stability in judgment when processing texts with similar semantics but different social orientations. It significantly reduces the frequency of conflicts between bias judgments and semantic content. This finding confirms the importance of interpretability mechanisms in implicit bias detection tasks. Compared with traditional methods based on classification or scoring, the approach in this paper emphasizes the mapping between internal model behavior and output semantics. It helps reveal the pathways and tendencies of implicit stereotype formation. This lays the groundwork for safety assessment and bias intervention in language models.

In addition, this study provides a more refined technical perspective on the social value and ethical risks of large language models. It offers practical guidance in fields such as education, justice, healthcare, and recruitment, where fairness in language generation is critically important. By improving the model's ability to detect fine-grained bias, the proposed framework supports the development of more trustworthy, transparent, and controllable natural language processing systems. It also provides a model foundation for the standardized application of AI systems in public-facing services.

Future work may expand the method's adaptability to cross-cultural and multilingual contexts. It can also explore the integration of external knowledge bases and user feedback signals to enable more flexible and generalizable bias detection and explanation platforms. Furthermore, embedding interpretability mechanisms into model interaction interfaces could allow users to perceive and respond to potential biases. This would support the responsible deployment and broader application of large language models in real-world social scenarios.

REFERENCES

[1] Y. Xiao, et al., "Fairness Mediator: Neutralize stereotype associations to mitigate bias in large language models," arXiv preprint arXiv:2504.07787, 2025.

[2] H. Kotek, R. Dockum, and D. Sun, "Gender bias and stereotypes in large language models," Proceedings of the ACM Collective Intelligence Conference, 2023.

[3] B. J. Chen, S. Binnewies, and B. Stantic, "Prompts de-biasing augmentation to mitigate gender stereotypes in large language models," Proceedings of the Asian Conference on Intelligent Information and Database Systems, Singapore: Springer Nature Singapore, 2025.

[4] R. Görge, M. Mock, and H. Allende-Cid, "Detecting linguistic indicators for stereotype assessment with large language models," arXiv preprint arXiv:2502.19160, 2025.

[5] K. S. Shejole and P. Bhattacharyya, "Detecting stereotypes and anti-stereotypes the correct way using social psychological underpinnings," arXiv preprint arXiv:2504.03352, 2025.

[6] Y. Bai, et al., "FairMonitor: A four-stage automatic framework for detecting stereotypes and biases in large language models," arXiv preprint arXiv:2308.10397, 2023.

[7] Z. Wu, S. Bulathwela, and A. Koshiyama, "Towards auditing large language models: Improving text-based stereotype detection," Socially Responsible Language Modelling Research, 2023.

[8] Y. Peng, Y. Wang, Z. Fang, L. Zhu, Y. Deng and Y. Duan, "Revisiting LoRA: A smarter low-rank approach for efficient model adaptation," *Proceedings of the 2025 5th International Conference on Artificial Intelligence and Industrial Technology Applications*, pp. 1248-1252, 2025.

[9] X. Meng, Y. Wu, Y. Tian, X. Hu, T. Kang and J. Du, "Collaborative distillation strategies for parameter-efficient language model deployment," *arXiv preprint* arXiv:2507.15198, 2025.

[10] Y. Wang, "Structured compression of large language models with sensitivity-aware pruning mechanisms," 2024.

[11] S. Wang, Y. Zhuang, R. Zhang and Z. Song, "Capsule network-based semantic intent modeling for human-computer interaction," *arXiv preprint* arXiv:2507.00540, 2025.

[12] X. Wang, "Time-aware and multi-source feature fusion for transformer-based medical text analysis," *Transactions on Computational and Scientific Methods*, vol. 4, no. 7, 2024.

[13] Y. Xing, "Bootstrapped structural prompting for analogical reasoning in pretrained language models," 2024.

[14] Y. Ma, G. Cai, F. Guo, Z. Fang and X. Wang, "Knowledge-informed policy structuring for multi-agent collaboration using language models," *Journal of Computer Science and Software Applications*, vol. 5, no. 5, 2025.

[15] X. Liu, Y. Qin, Q. Xu, Z. Liu, X. Guo and W. Xu, "Integrating knowledge graph reasoning with pretrained language models for structured anomaly detection," 2025.

[16] Q. Wu, "Task-aware structural reconfiguration for parameter-efficient fine-tuning of LLMs," 2024.

[17] H. Zheng, Y. Wang, R. Pan, G. Liu, B. Zhu and H. Zhang, "Structured gradient guidance for few-shot adaptation in large language models," *arXiv preprint* arXiv:2506.00726, 2025.

[18] W. Zhang, Z. Xu, Y. Tian, Y. Wu, M. Wang and X. Meng, "Unified instruction encoding and gradient coordination for multi-task language models," 2025.

[19] X. Han, Y. Sun, W. Huang, H. Zheng and J. Du, "Towards robust few-shot text classification using transformer architectures and dual loss strategies," *arXiv preprint* arXiv:2505.06145, 2025.

[20] Y. Zhao, W. Zhang, Y. Cheng, Z. Xu, Y. Tian and Z. Wei, "Entity boundary detection in social texts using BiLSTM-CRF with integrated social features," 2025.

[21] B. Pang, "Investigating stereotypical bias in large language and vision-language models," Dissertation, University of Auckland, New Zealand, 2025.

[22] R. Pan, "Policy-guided path selection and evaluation in multi-step reasoning with large language models," Transactions on Computational and Scientific Methods, vol. 4, no. 8, 2024.

[23] Y. Peng, "Context-aligned and evidence-based detection of hallucinations in large language model outputs," 2025.

[24] X. Quan, "Layer-wise structural mapping for efficient domain transfer in language model distillation," Transactions on Computational and Scientific Methods, vol. 4, no. 5, 2024.

[25] W. Zhu, "Fast adaptation pipeline for LLMs through structured gradient approximation," 2024.

[26] C. May, et al., "On measuring social biases in sentence encoders," arXiv preprint arXiv:1903.10561, 2019.

[27] M. Nadeem, A. Bethke, and S. Reddy, "StereoSet: Measuring stereotypical bias in pretrained language models," arXiv preprint arXiv:2004.09456, 2020.

[28] H. Zhou, et al., "Unibias: Unveiling and mitigating LLM bias through internal attention and FFN manipulation," arXiv preprint arXiv:2405.20612, 2024.

[29] Y. Hong and E. Yang, "Unbiased classification through bias-contrastive and bias-balanced learning," Advances in Neural Information Processing Systems, vol. 34, pp. 26449-26461, 2021.